\documentclass{article}

\usepackage[dblblindworkshop, final]{neurips_2025}
\vspace{-1em}
\workshoptitle{Generative AI in Finance}

\usepackage[utf8]{inputenc} 
\usepackage[T1]{fontenc}    
\usepackage{hyperref}       
\usepackage{url}            
\usepackage{booktabs}       
\usepackage{amsfonts}       
\usepackage{nicefrac}       
\usepackage{microtype}      
\usepackage{xcolor}         
\usepackage{graphicx}
\usepackage{amsmath}

\usepackage{float}
\usepackage{wrapfig}
\usepackage{placeins}
\usepackage{footnote}
\usepackage{listings}
\usepackage{bbm}
\usepackage{subcaption}
\title{A Fast and Effective Solution to the Problem of Look-ahead Bias in LLMs}

\author{%
  Humzah Merchant \\
  University of Chicago\\
  Chicago, IL 60637 \\
  \And
  Bradford Levy\thanks{Direct correspondence to \texttt{bll@uchicago.edu}. \href{https://github.com/fin-ai-lab/fast-effective-solution-to-lab}{Replication code and data.} \href{https://frontiertopit.com}{\textcolor{blue}{\textbf{Visit frontiertopit.com} $\rightarrow$}}} \\
  University of Chicago \\
  Chicago, IL 60637 \\
}

\begin{document}

\maketitle

  \vspace{-30pt}
  \begin{center}
        \small
        \href{https://frontiertopit.com}{\textcolor{blue}{\textbf{Visit frontiertopit.com} $\rightarrow$}}
\end{center}
    \vspace{-5pt}

\begin{abstract}
Applying LLMs to predictive tasks in finance is challenging due to look-ahead bias resulting from their training on long time-series data. This precludes the backtests typically employed in finance since retraining frontier models from scratch with a specific knowledge cutoff is prohibitive. In this paper, we introduce a fast, effective, and low-cost alternative. Our method guides generation at inference time by adjusting the logits of a large base model using a pair of smaller, specialized models—one fine-tuned on information to be forgotten and another on information to be retained. We demonstrate that our method effectively removes both verbatim and semantic knowledge, corrects biases, and outperforms prior methods.
\end{abstract}

\section{Introduction}
LLMs have demonstrated remarkable capabilities, achieving state-of-the-art performance on a diverse array of natural language tasks \citep{brown2020languagemodelsfewshotlearners, touvron2023llama2openfoundation, DeepSeekAI2024} and specialized financial applications \citep{chen_exp_ret_llms, lopezlira2024chatgptforecaststockprice, wang2024beancounter}. However, applying LLMs to the predictive tasks common in financial applications is challenging due to the fact that they are trained on long time-series \citep{longpre-etal-2024-pretrainers} and known to memorize content \citep{Carlini2021ExtractingTD, carlini2023quantifying}. For example, if an LLM has memorized historical earnings figures and stock prices, then estimates of predictive ability based on historical data, i.e., a backtest, are likely overly optimistic \citep[see, e.g.,][]{glasserman2023assessinglookaheadbiasstock, levy_Caution, lopezlira2024chatgptforecaststockprice}.

The simplest solution to such concerns is training a model from scratch with a given historical knowledge cutoff \citep[e.g.,][]{drinkall-etal-2024-time}. While simple, the very scale that enables powerful generalization of LLMs creates significant challenges to doing this. First, it is likely cost prohibitive to train a frontier quality LLM purely to enable backtesting \citep{hoffmann2022trainingcomputeoptimallargelanguage, grattafiori2024, deepseekai2025deepseekv3technicalreport}. Second, even if budget allowed, modern foundation models are trained on trillions of tokens, much of which collected only recently. Thus, data availability alone may preclude training a ``historical'' frontier model.

In this work, we propose a theoretically motivated and conceptually simple alternative to training from scratch: \textbf{Inference-Time Unlearning}. Rather than modifying the weights of a frontier model, our method guides generation at inference time by using a pair of smaller, specialized models. One small model is fine-tuned on the data to be forgotten, e.g., knowledge after 2010, while another is tuned on a proxy for the data to be retained, e.g., knowledge prior to 2010. By modifying the logits of the frontier model with the difference of the specialized models, through a variety of experiments we document that our method effectively unlearns unwanted content while preserving model utility.

Our paper makes three primary contributions to the literature on machine unlearning:
\begin{enumerate}
    \item \textbf{Efficacy}: We demonstrate that our method (i) effectively removes both verbatim and semantic knowledge from a model, and (ii) can correct unwanted biases such as primacy.
    \item \textbf{Efficiency}: By restricting fine-tuning to small auxiliary models (with orders of magnitude fewer parameters than the base LLM), our approach drastically reduces the computational cost of unlearning. For example, we find that even simple tri-gram based LMs are effective. This makes on-demand unlearning practical and scalable.
    \item \textbf{Utility Preservation}: Our method maintains the model's performance on general knowledge and standard evaluation benchmarks. Because the base model's weights remain unchanged, the impact on its core capabilities is minimal, outperforming prior methods in preserving utility as the number of unlearning requests grows.
\end{enumerate}

Collectively, we demonstrate that our approach provides a practical, low-cost, and effective solution to the critical problem of selectively forgetting information in LLMs, paving the way for more reliable evaluation of their abilities within financial applications.

\section{Method}
We begin by defining the problem, introducing our method, and finally connecting it to existing work. Let $V$ denote a finite vocabulary of tokens. A token sequence of length $T$ is denoted as $x = (x_1, x_2, ..., x_T)$ where each token $x_t \in V$. The prefix of a token sequence up to token $t-1$ is denoted $x_{<t}=(x_1, ..., x_{t-1})$. There are two data generating distributions $D_A$ and $D_B$ where the support of $D_B$ is contained within $D_A$. Finally, $P(x_t|x_{<t})$ and $Q(x_t|x_{<t})$ denote the conditional token distributions under $D_A$ and $D_B$, respectively.

We consider the situation where we wish to sample from $Q$ but do not have access to it. Instead, only $P$ is accessible. For example, $P$ could be a large frontier model for which it is cost prohibitive to retrain a new model from scratch on $D_B$. Within the finance domain, $Q$ could be a model as capable as $P$ but trained up to a fixed knowledge cutoff so as to avoid look-ahead bias. Generally, our goal is to approximate sampling from $Q$ using only $P$ and samples drawn from $D_A$ and $D_B$.

\subsection{Divergence decoding}
Consider two small models $p(x_t|x_{<t})$ and $q(x_t|x_{<t})$ trained on samples from $D_A$ and $D_B$, respectively. Denote the logits of a given model $M$ as $l_M(x_{<t}) \in \mathbb{R}^{|V|}$. Divergence Decoding (DD) approximates sampling from $Q$ by adjusting the logits of $P$ according to the divergence between $q$ and $p$. Empirically, we consider two adjustments. The first is a linear combination of the logits,
\begin{align}
    \hat{l}^{LC}_Q(x_{<t}) &= l_P(x_{<t}) + \alpha \cdot [l_q(x_{<t}) - l_p(x_{<t})], \label{eq:dd_logits}
\end{align}

while the second adjustment is rank based,
\begin{align}
    \hat{l}^{R}_Q(x_{<t}) &= l_P(x_{<t}) - \mathbbm{1}_{rank(l_p(x_{<t}) - l_q(x_{<t})) \leq k} \cdot \infty. \label{eq:dd_logits_rank}
\end{align}

In the case of the linear adjustment, if the difference between $Q$ and $P$ is indeed linear in logit space, then there exists some value of $\alpha$, $p$, and $q$ which enables $Q$ to be perfectly recovered. If the difference is not linear however, then this is not true. For this reason, we also explore the rank based approach, which prevents generating the top-$k$ most divergent tokens between $p$ and $q$.

Samples can then be drawn via typical methods \citep[e.g.,][]{fan-etal-2018-hierarchical,Holtzman2020The} from the approximation,
\begin{align}
    \widehat{Q}(x_t|x_{<t}) &= \text{softmax}(\hat{l}_Q(x_{<t})). \label{eq:dd_samples}
\end{align}
While the adjustments in Eq. \ref{eq:dd_logits} and \ref{eq:dd_logits_rank} require additional forward passes for $p$ and $q$, we show in Section \ref{sec:experiments} that strong performance can be achieved even when $p$ and $q$ are trigram models---which add negligible computational overhead (see Figure \ref{fig:muse_model_size}).

\subsection{Theoretical motivation}
While simple to implement and fast at inference time, our method is theoretically motivated by the Product of Experts \citep{Hinton1999ProductsOE} and Importance Sampling \citep{hammersley1965monte} literature. In Appendix \ref{app:theory:PoE}, we show that the approximation $\widehat{Q}$ can be formulated as a Product of Experts model,

\begin{equation}
    \widehat{Q}(x_t|x_{<t}) \propto \underbrace{\vphantom{ \Bigg(\frac{a}{b}\Bigg)^{0.3} }P(x_t|x_{<t})}_\text{Base Expert} \cdot \underbrace{\Bigg[\frac{q(x_t|x_{<t})}{p(x_t|x_{<t})}\Bigg]^\alpha}_\text{Domain Expert}
\end{equation}

where $\widehat{Q}$ is the product of a ``Base Expert'' $P$ responsible for providing foundational knowledge and a ``Domain Expert'' comprised of the ratio of $q$ to $p$. Intuitively, the role of the domain expert can be summarized by three cases:
\begin{enumerate}
    \item $q \approx p$: Tokens are similarly likely under both $D_A$ and $D_B$ and the domain expert ratio is close to 1 effectively leaving the probabilities from the base model $P$ unchanged
    \item $q \gg p$: Tokens are much \textbf{\textit{more}} likely under $D_B$ than $D_A$, and the domain expert ``upvotes'' such tokens by \textbf{\textit{increasing}} the probability assigned to them
    \item $q \ll p$: Tokens are much \textbf{\textit{less}} likely under $D_B$ than $D_A$, and the domain expert ``downvotes'' such tokens by \textbf{\textit{decreasing}} the probability assigned to them
\end{enumerate}

Finally, DD can also be linked to importance sampling in Monte Carlo analysis whereby the expectation of some function $f(x)$ under a target distribution $D_{target}$ is estimated using samples drawn from a proposal $D_{proposal}$. Formally,

\begin{equation}
    \mathbb{E}_{x \sim D_{target}}[f(x)] = \mathbb{E}_{x \sim D_{proposal}}\bigg[f(x) \frac{D_{target}(x)}{D_{proposal}(x)}\bigg],
\end{equation}

where the importance weight $w(x) = \frac{D_{target}(x)}{D_{proposal}(x)}$ adjusts the expectation taken over $D_{proposal}$ for differences between the proposal and target distributions. Analogously, divergence decoding uses the ratio of $q$ to $p$ to adjust for differences between the inaccessible model $Q$ and accessible one $P$.

\section{Experiments}
\label{sec:experiments} 
Next, we explore the efficacy of our method across a variety of empirical settings. First, we evaluate its performance on the well-established benchmark ``MUSE: Machine Unlearning Six-Way Evaluation for Language Models'' \citep{shi2025muse}. We then explore performance on finance specific tasks: (i) unlearning knowledge of mergers and acquisitions, and (ii) debiasing forecasts of future performance.
\subsection{MUSE unlearning benchmark}
\begin{figure}[H]
    \centering
    \includegraphics[width=1\linewidth]{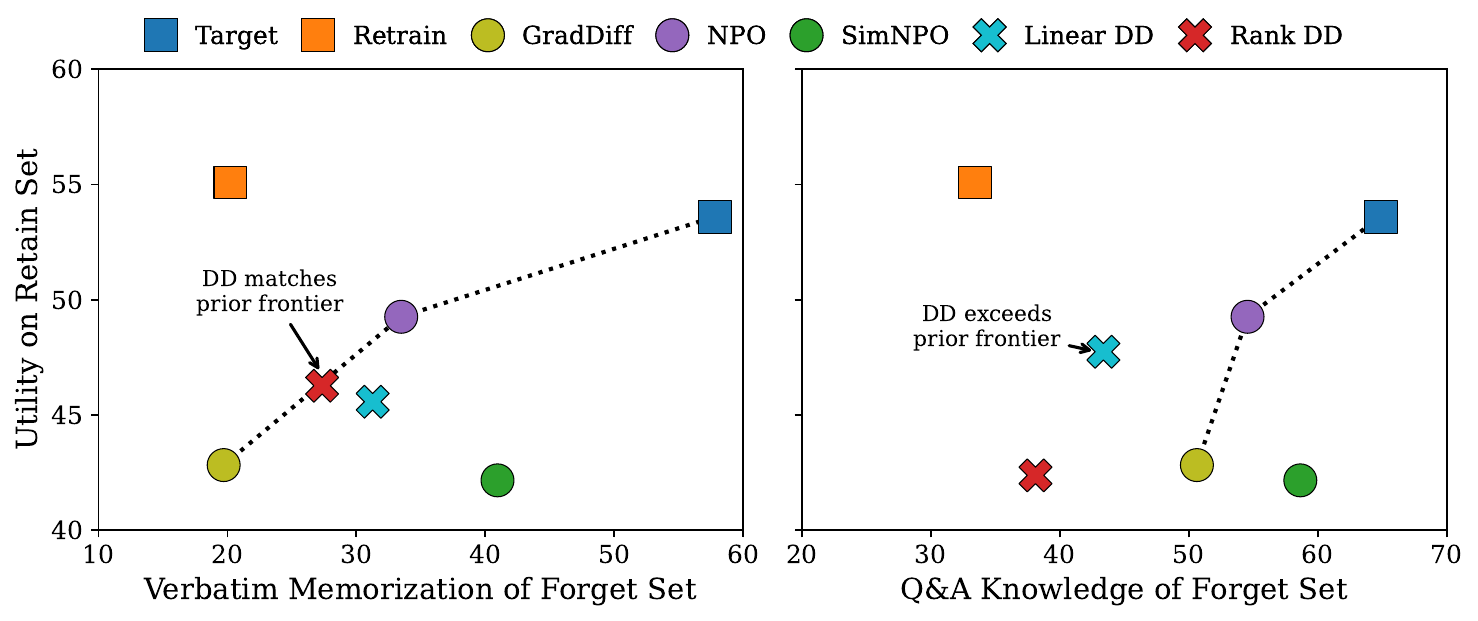}
    \caption{MUSE Results: Target is the model to which unlearning is applied. Retrain is the best---but most costly---result of retraining from scratch. Closer to Retrain is better.}
    \label{fig:muse_results}
\end{figure}

We apply our method to the news dataset from the MUSE benchmark \citep{shi2023detecting, shi2025muse}. For our small specialized models, $p$ and $q$, we use the \textit{princeton-nlp/Sheared-LLaMA-1.3B} \citep{xia2023sheared} GPT model---because it shares a tokenizer with the benchmark models used by MUSE \citep{shi2025muse}---as well as trigram LMs based on \textit{Stupid Backoff} \citep{brants-etal-2007-large}. Across both types of models, we find that our method matches or pushes the frontier of unlearning while preserving model utility (see Figure \ref{fig:muse_results}). This is particularly notable given the reduced computational costs of our method relative to prior work (see Section \ref{appendix-computational-analyses}) e.g., trigram based models are effectively costless (see Figure \ref{fig:muse_model_size}). In additional tests we evaluate strong performance on the \textbf{privacy}, \textbf{scalability}, and \textbf{sustainability} benchmarks (see Section \ref{sec:appendix-MUSE}).

\subsection{Finance specific tasks}\label{sec:applied-experiments}
While the MUSE results suggest that divergence decoding is effective and efficient, MUSE is a general benchmark. In this section, we seek to apply our method to finance focused tasks. Throughout our experiments, we use the \textit{google/gemma-3-27b-it} and \textit{google/gemma-3-4b-it} models \citep{gemma_2025}. When fine-tuning the 4B models, we found that the instruction following was brittle. To alleviate this issue, we trained using samples in question and answer format generated from the larger model such as \textit{Who did \{acquirer\} acquire in \{year\}?} See Section~\ref{sec:appendix-experiments} for detailed experimental setups.

\subsubsection{Unlearning mergers and acquisitions}

\begin{figure}[h]
    \centering
    \begin{subfigure}{0.49\textwidth}
        \centering
        \includegraphics[width=\linewidth]{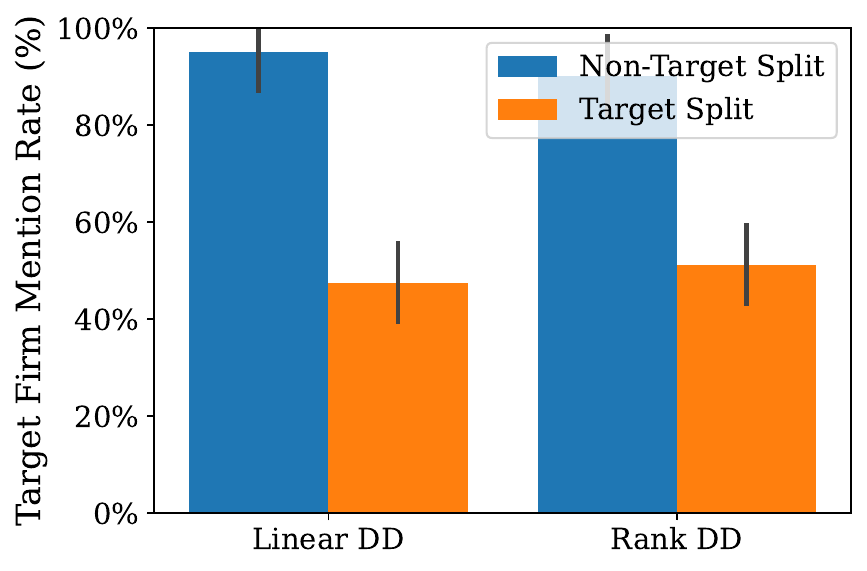}
        \caption{Effective targeted unlearning with minimal utility loss in a complex, instruction tuned environment.}
        \label{fig:ma_unlearning}
    \end{subfigure}
    \hfill
    \begin{subfigure}{0.49\textwidth}
        \centering
        \includegraphics[width=\linewidth]{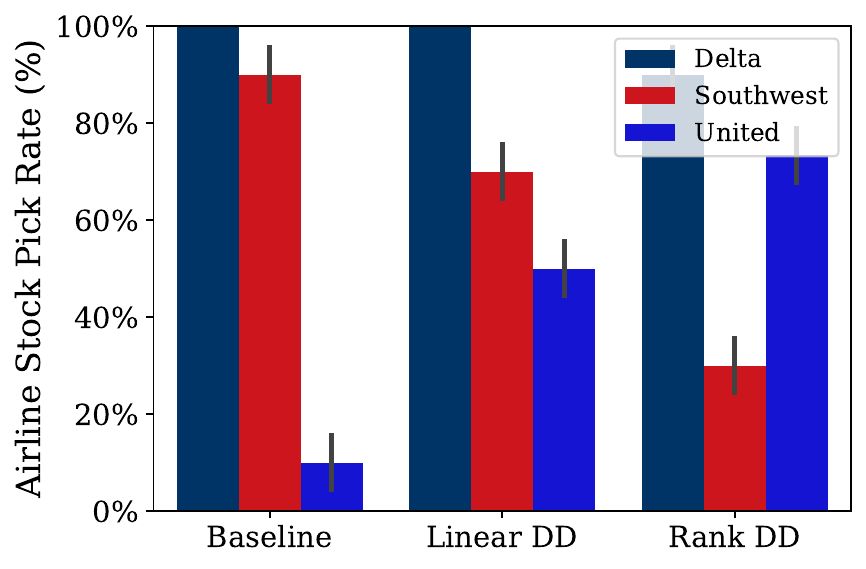}
        \caption{Unlearning now irrelevant historical performance leads to change in portfolio choices}
        \label{fig:llm_debiasing}
    \end{subfigure}
    \caption{Performance on finance specific tasks. 99\% confidence intervals presented where applicable.}
\end{figure}

We split M\&A deals that Gemma memorized into two balanced sets, rotating both as the target and non target splits. We then prompted Gemma with a DD setup to suggest a target firm for acquisition. Across both the linear and rank-based divergence decoding implementations we find a significant reduction in the memorization of M\&A deals (see Figure \ref{fig:ma_unlearning}). In Section \ref{sec:appendix-mmlu} we also apply the setups to MMLU CoT and see almost no loss in model performance. 

\subsubsection{Debiasing expectations of future performance}
Recent work has found that LLMs will associate long-run general sentiment with expectations of future performance even in the face of content which suggests a future reversal of sentiment \citep{glasserman2023assessinglookaheadbiasstock}. In our final experiment, we explore whether our method can remove such primacy biases from LLM-based forecasts. Specifically, we ask Gemma 27B to build stock portfolios, picking two stocks each for several industries. Figure \ref{fig:llm_debiasing} presents results for the airline industry---where Southwest and United have had such reversals. We finetuned the `forget' small model on reports of both airlines' performances in 2014, 2015, and 2016, and finetuned the `retain' small model on reports of 2022, 2023, and 2024.  We find that DD successfully reduces this bias.

\section{Conclusion}
In this paper, we introduce a novel inference-time unlearning method, ``Divergence Decoding.'' Instead of costly retraining, this approach guides a large model's output by leveraging two smaller, specialized models—one fine-tuned on data to "forget" and another on data to "retain"—without altering the base model's weights. By adjusting the large model's logits based on the difference between these auxiliary models, the system effectively unlearns content---or prevents the use of future knowledge in financial applications. Across several experiments, we confirm the method is highly effective at removing targeted knowledge and correcting biases, computationally efficient (particularly with simple trigram models), and successfully preserves the base model's general utility. This provides a practical, low-cost solution for selective forgetting, enabling more reliable evaluation of LLMs in chronologically sensitive domains like finance.

\begin{ack}
We thank Ralph Koijen, Sanjog Misra, and Ted Sumers for helpful discussions. We acknowledge generous financial support from the Booth School of Business and the Center for
Applied AI. The authors have no competing interests to disclose.
\end{ack}

\par
\bibliographystyle{plain} 
\bibliography{references} 


\appendix
\section{Analyses}
\label{appendix-theory}
\subsection{Divergence Decoding and Product of Experts}\label{app:theory:PoE}
\cite{Hinton1999ProductsOE} introduced the Product of Experts (PoE) framework whereby $n$ probability models are multiplicatively combined into a single model. Let the $i$-th expert be denoted by $f_i(x|\theta_i)$, then a PoE model $R$ comprised of $n$ experts is given by,
\begin{align}
    R(x|\theta_1, ..., \theta_n) &= \frac{1}{Z}\prod_{i=1}^n f_i(x|\theta_i),
\end{align}
where $Z$ is a normalization constant. To highlight the connection between divergence decoding and PoE, recall Eq. \ref{eq:dd_logits}:
\begin{equation*}
    \hat{l}_Q(x_{<t}) = l_P(x_{<t}) + \alpha \cdot [l_q(x_{<t}) - l_p(x_{<t})].
\end{equation*}
In Eq. \ref{eq:dd_logits}, a given model $M$ has logits which are equal to the log-probabilities up to an additive constant which depends on the token sequence prefix $x_{<t}$ but not the token $x_t$, i.e.,
\begin{equation}
    l_M(x_{<t}) = \log M(x_t|x_{<t}) + C_M(x_{<t}). \label{eq:app_logits_log_probs}
\end{equation}
Substituting Eq. \ref{eq:app_logits_log_probs} into Eq. \ref{eq:dd_logits} for each model, gathering the constants, and performing some algebra reveals the link to PoE:
\begin{align*}
    \log\widehat{Q}(x_t|x_{<t}) &= \log P(x_t|x_{<t}) + \alpha \cdot [\log q(x_t|x_{<t}) - \log p(x_t|x_{<t})] + C \\
    \widehat{Q}(x_t|x_{<t}) &\propto \exp\big( \log P(x_t|x_{<t}) + \alpha \cdot [\log q(x_t|x_{<t}) - \log p(x_t|x_{<t})] \big) \\
    &\propto P(x_t|x_{<t}) \cdot q(x_t|x_{<t})^\alpha \cdot p(x_t|x_{<t})] ^{-\alpha} \\
    &\propto P(x_t|x_{<t}) \cdot \Bigg[\frac{q(x_t|x_{<t})}{p(x_t|x_{<t})}\Bigg]^\alpha.
\end{align*}

\subsection{Computational Analyses}
\label{appendix-computational-analyses}

We want to analyze the \textbf{inference-time cost} due to running the small models in tandem with the large model. Let $N$ denote the number of parameters in the large model and $n$ the number of parameters in each small model.  
Measured in FLOPs \cite{kaplan2020scalinglawsneurallanguage}, the inference cost scales from

\[
2N \;\longrightarrow\; 2(N+2n).
\]

Additionally, let $d_r$ and $d_f$ be the sizes of the retain and forget datasets (in tokens), let $e_N$ and $e_n$ be the number of epochs the large and small models are trained for, respectively, and let $I$ be the number of inference tokens. Hence, we want to know after how many inference tokens does it become more costly to use DD over another method, \textbf{assuming both work equally well}. Considering one of the simplest unlearning methods, Gradient Ascent \citep{jang2023knowledge} \textbf{without any kind of regularizer}, DD becomes more costly once:

\[
\begin{aligned}
6ne_n(d_r + d_f) + 2(N+2n)I &\geq 6Ne_N(d_f) + 2NI \\
I &\geq \frac{3Ne_N d_f}{2n} - \frac{3e_n(d_r + d_f)}{2}
\end{aligned}
\]

\textbf{For many financial applications, such as backtesting,} $d_f \gg I$ by many magnitudes. 

\subsection{Limitation - Instruction Tuning Sensitivity}
\label{appendix:limitations}

A key limitation lies in the method’s \textbf{sensitivity to instruction-tuning}. For instance, when unlearning financial knowledge, the model may generate stock recommendations in the format:

\begin{center}
``**1. \{firm name\}:**"
\end{center}

If the smaller models anticipate a different structure (e.g., a ticker symbol or bullet marker after the `1.'), the divergence in logits at the critical step may be diluted or entirely noisy. Worse, if one small model aligns closely with the large model while the other does not, differences fail to cancel and can yield unstable or unintended outputs. Therefore, researchers adopting this method may therefore need to carefully re-tune instruction following behavior.

\section{Additional Results and Experimental Setups}

Everything was run on an H100 cluster unless otherwise noted. Code and data is available at \href{https://github.com/fin-ai-lab/fast-effective-solution-to-lab}{GitHub}.

\label{sec:appendix-experiments}
\subsection{MUSE}
\label{sec:appendix-MUSE}

We finetune the LlaMA models using the \textbf{AdamW Torch optimizer} and a \textbf{cosine scheduler} for \textbf{10} epochs. We set the learning rate such that the loss approximately halves over the course of training. 

We sweep $\alpha \in \{0.5, 0.6,\dots, 1.5\}$ and $\text{top-}k \in \{1, 5, 20, 50, 100, 200, 500, 1000\}$ for the LLaMA models, and at $\alpha \in \{5, 10,\dots30\}$ and $\text{top-}k \in \{1, 2, 3, 5, 10\}$ for the trigram models. We choose the most optimal point as the point closest in euclidean distance to Retrain. We find that in general, rank DD outperforms on verbatim memorization while linear DD outperforms on Q\&A knowledge.

\begin{table}[H]
    \centering
    \caption{Configuration MUSE}
    \begin{tabular}{lrrrr}
        Model & Initial LR & Best Verbatim & Best Q\&A \\
        \toprule
         Stupid Backoff Trigram & & TopK=1 & Alpha=10 \\
         princeton-nlp/Sheared-LLaMA-1.3B & 5e-5 & TopK=100 & Alpha=0.8 \\
         princeton-nlp/Sheared-LLaMA-2.7B & 4e-5 & TopK=200 & Alpha=1.0 \\
    \end{tabular}
    \label{tab:placeholder}
    \vspace{-20pt}
\end{table}

\begin{figure}[H]
    \centering
    \includegraphics[width=1\linewidth]{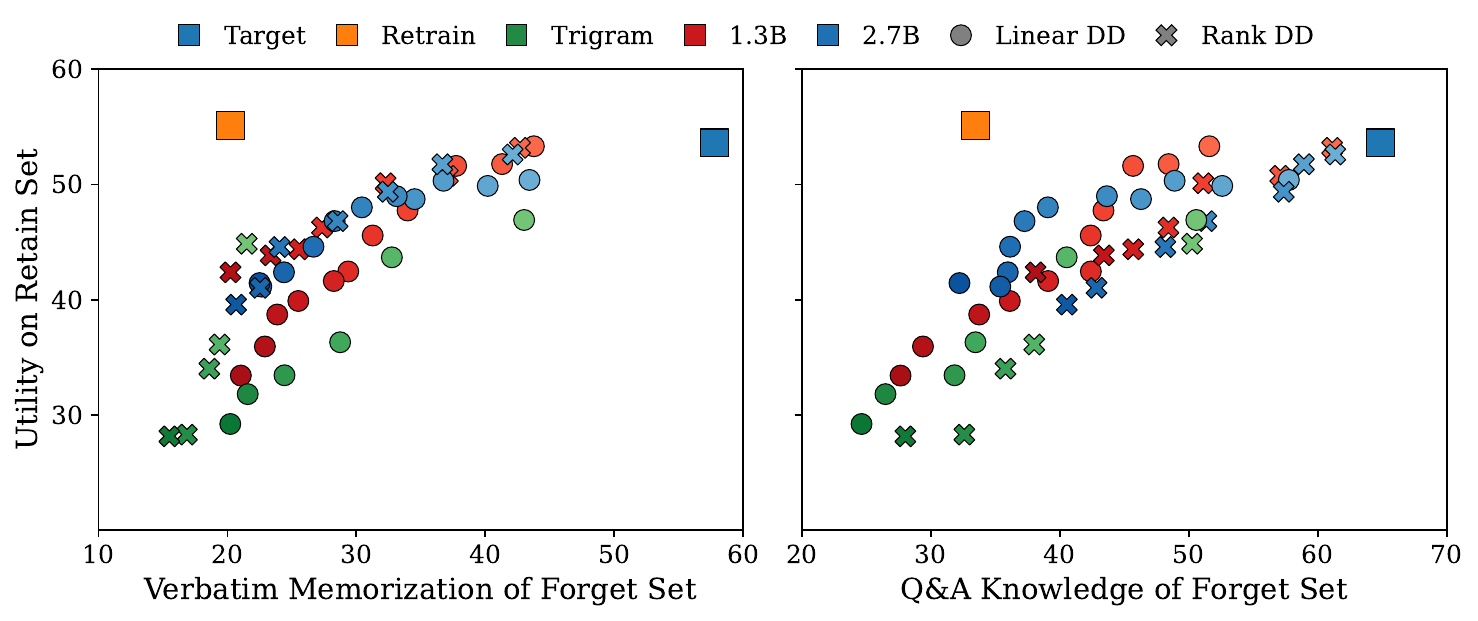}
    \caption{All hyper-parameter and model size configurations}
    \label{fig:placeholder}
\end{figure}

For the other methods, we use the default settings provided by OpenUnlearning

\begin{table}[H]
    \centering
    \caption{MUSE Configurations}
    \begin{tabular}{lll}
        \toprule
        Method & Epochs & Method-Specific Hyperparameters \\
        \midrule
        GradDiff & 1$^{*}$ & $\alpha=1.0, \ \gamma=1.0$ \\
        NPO      & 10      & $\beta=0.1, \ \alpha=1.0, \ \gamma=1.0$ \\
        SimNPO   & 10      & $\delta=0, \ \beta=4.5, \ \alpha=1.0, \ \gamma=0.125$ \\
        \bottomrule
    \end{tabular}
    \label{tab:muse_configs}
    \vspace{0.5em}
    \begin{minipage}{\linewidth}
        \footnotesize
        \raggedright
        Default hyperparameters: batch size = 32, learning rate = $1\!\times\!10^{-5}$, warmup epochs = 1, weight decay = 0.01, retain loss = NLL. $^{*}$ For GradDiff, the 1 epoch setting is the only deviation from the defaults.
    \end{minipage}
\end{table}

\subsubsection{Model size}
Given that applying our method using the 1.3B models for $p$ and $q$ is effective, a natural question is how sensitive this performance is to model size. We investigate this using \textit{princeton-nlp/Sheared-LLaMA-2.7B} and trigram LMs based on \textit{Stupid Backoff} \citep{brants-etal-2007-large}. We select the most optimal configuration of every model size, based on the minimum euclidean distance to Retrain, and rescale the metric such that Target is 100\%. The Trigram models outperform on the Verbatim Memorization and perform slightly worse than the LLMs on Q\&A. Upon further inspection of the Q\&A questions where the Trigram models perform well, we find that this is largely due to questions which are more similar to the underlying training data. Thus, we conclude that the Trigram models are likely most useful for unlearning verbatim content.

\begin{figure}[h]
    \centering
    \begin{subfigure}{0.49\textwidth}
        \centering
        \includegraphics[width=\linewidth]{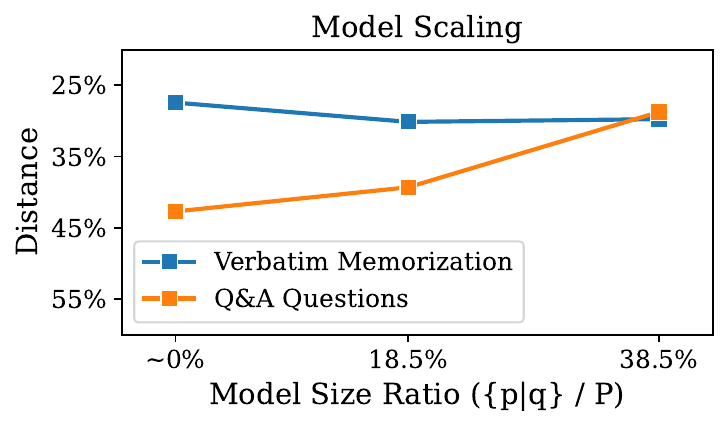}
        \caption{Verbatim memorization favors smaller $p,q$; Q\&A favors larger $p,q$.}
        \label{fig:muse_model_size}
    \end{subfigure}
    \hfill
    \begin{subfigure}{0.49\textwidth}
        \centering
        \includegraphics[width=\linewidth]{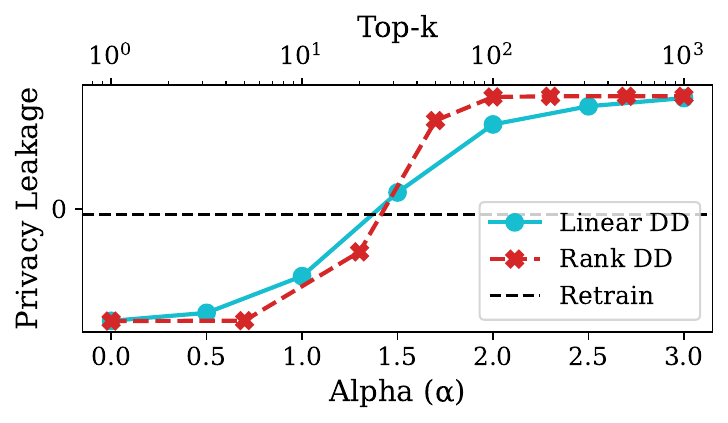}
        \caption{A broad range of hyper-parameter settings balance over- and under-unlearning.}
        \label{fig:muse_priv}
    \end{subfigure}
    \caption{Analysis of Model Scaling and Over- or Under- Unlearning on MUSE}
\end{figure}

\subsubsection{Sustainability and Scaling}
Finally, prior work has found that many unlearning methods exhibit poor scalability---the unlearning of very large amounts of content---and sustainability---sequential requests to unlearn additional content. We explore the efficacy of our method along these dimensions using the MUSE scaling and sustainability benchmarks to ensure that performance does not degrade. To extend the benchmark, we additionally measure performance on the original forget set (Q\&A), ensuring that improved generalization \textbf{does not} come at the cost of overwriting prior forgetting, specifically with the weights of the forget model being overwritten. 

\begin{figure}[H]
    \centering
    \includegraphics[width=1\linewidth]{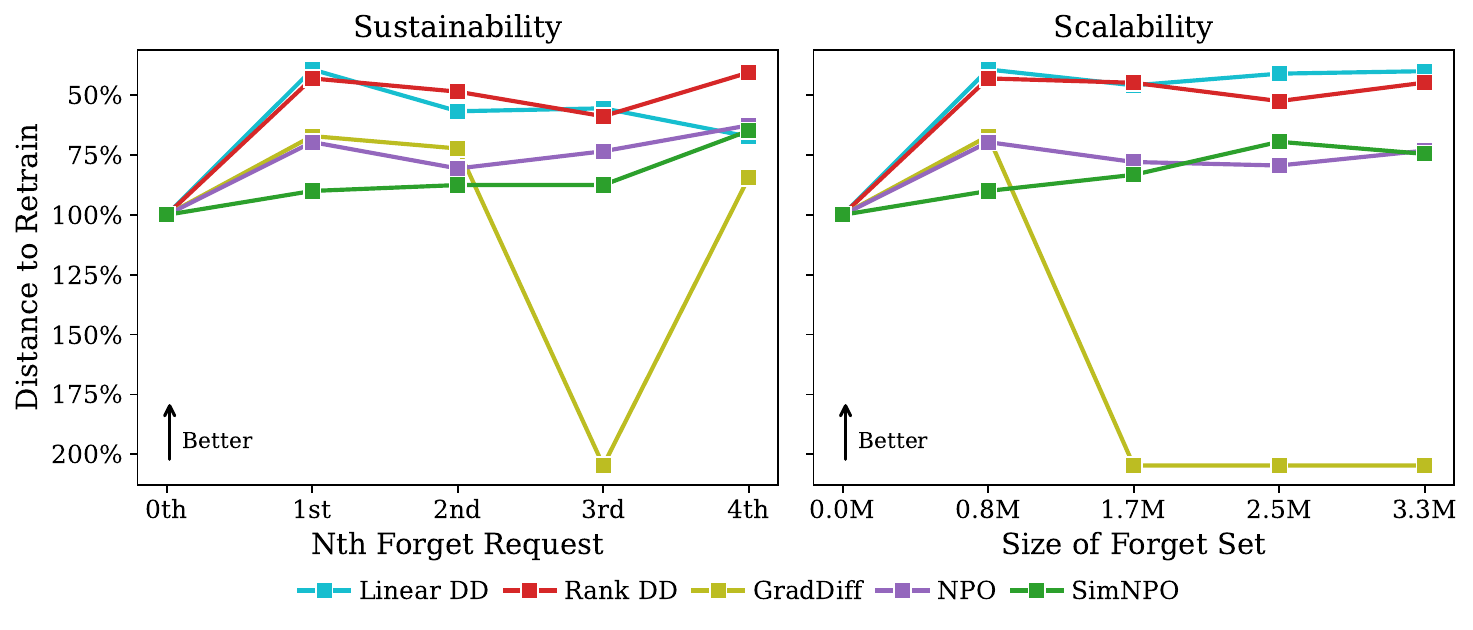}
    \caption{MUSE Scaling and Sustainability. The left column is sustainability - consecutive forget sets of the same size - and the right column is scaling, increasingly large forget sets. We evaluate both utility on the retain set and utility on the \textbf{original} forget set, in order to ensure that we are not losing forget ability, and take the euclidean distance to Retrain with Target as 100\%.}
    \label{fig:scal_and_sust}
\end{figure}

\subsection{MMLU}
\label{sec:appendix-mmlu}

\begin{table}[H]
  \caption{MMLU CoT 0-Shot}
  \vspace{7pt}
  \label{mmlu-cot}
  \centering
  \begin{tabular}{ll}
    \toprule
    Name & Score \\
    \midrule
    google/gemma-3-27b-it & 77.94 \\
    Unlearning Split B $\alpha$=2  & 77.58 \\
    Unlearning Split B TopK=250 & 75.09 \\
    \bottomrule
  \end{tabular}
\end{table}

We use the Language Model Evaluation Harness \citep{eval-harness} by EleutherAI and modify the template from \href{https://github.com/EleutherAI/lm-evaluation-harness/blob/84aa9f95fea2e1bd298e1859cab0b12094f80e0b/lm_eval/tasks/gsm8k/gsm8k-cot-llama.yaml#L81}{gsm8k-cot-llama}. We run some of our evaluations on an A40 and others on an H100, all with 8-bit quantization and greedy decoding (temperature of 0).

\subsection{Applied Experiments}
For both experiments, we finetuned the models using the \textbf{AdamW Torch optimizer}, a \textbf{cosine scheduler}, a starting learning rate of \textbf{1e-5}, for \textbf{3 epochs}. We would pass the full question and answer but mask any tokens outside the main response from the loss calculation to prevent the model from learning the question. When tuning our training parameters (we did M\&A first and repeated the settings,) we would randomly select one prompt per deal to use as a simple validation set. However, our final models were trained on the full dataset. \label{appendix:standard-errors} We calculated standard errors by de-meaning all groups and calculating a population standard error.

\subsubsection{M\&A}
We sample deals from Wharton Research Data Services, SDC - Mergers and Acquisitions, across the whole database, from 2010-01-01 to 2024-08-01, using the following filters to get \textbf{294} high profile conventional M\&A deals. 
\begin{lstlisting}
HERE form = 'Merger' AND 
apublic = 'Public' AND 
anation = 'United States' AND 
afinancial = 'No' AND 
albofirm = 'No' AND 
alp = 'No' AND 
status = 'Completed' AND 
deal\_value >= 5000
\end{lstlisting}

We do a first pass filtering, running `\textit{What firms did \{acquirer name\} acquire in \{year\}? List without explanation.}' four times for each sample for each model and keeping any samples that Gemma 27b mentions the target at least 3/4 times and Gemma 4b mentions the target no more then 1/4 times. We want to throw away deals that Gemma 4b knows because if Gemma 4b already knows a deal then finetuning will have a smaller impact on the final logit distribution.

We then apply our second stage and final unlearning prompt 
\begin{lstlisting}
It's the end of {year-1}. What two or three companies do you think {acquirer name} might most consider acquiring in {year}? 
\end{lstlisting}
and split the remaining 41 deals into two sets by selecting alternating deals chronologically.

\begin{table}[H]
  \caption{Final Deal Sets}
  \label{deals}
  \centering
  \scriptsize
  \begin{tabular}{cc}
    \toprule
    Set A & Set B \\
    \midrule
    Bucyrus International <> Caterpillar & Allegheny Energy <> FirstEnergy \\
    El Paso <> Kinder Morgan & Goodrich <> United Technologies \\
    Hillshire Brands <> Tyson Foods & Biomet <> Zimmer Holdings \\
    Rockwood Holdings <> Albemarle & Family Dollar Stores <> Dollar Tree \\
    Rock-Tenn <> MeadWestva& Bally Technologies <> Scientific Games \\
    Bright House Networks LLC <> Charter Communications & Pharmacyclics <> AbbVie \\
    Time Warner Cable <> Charter Communications & Altera <> Intel \\
    Health Net <> Centene & Worldpay Group PLC <> Vantiv \\
    Alere <> Abbott Labs & Plum Creek Timber <> Weyerhaeuser \\
    Linear Technology <> Analog Devices & Medivation <> Pfizer \\
    Rice Energy <> EQT & NetSuite <> Oracle \\
    Kite Pharma <> Gilead Sciences & Scripps Networks Interactive <> Discovery Communications \\
    Arris International Plc <> Commscope Holding & DCT Industrial Trust <> Prologis \\
    Celgene <> Bristol-Myers Squibb & Tribune Media <> Nexstar Media Group \\
    Legg Mason <> Franklin Resources & SunTrust Banks <> BB\&T \\
    Immunomedics <> Gilead Sciences & Noble Energy <> Chevron \\
    Avast PLC <> NortonLifeLock & IHS Markit Ltd <> S\&P Global \\
    Activision Blizzard <> Microsoft & Cerner <> Oracle \\
    Duke Realty <> Prologis & Black Knight <> Intercontinental Exchange \\
    ABIOMED <> Johnson \& Johnson & VMware <> Broadcom \\
    & Oak Street Health <> CVS Health \\
    \bottomrule
  \end{tabular}
\end{table}

\begin{table}[H]
  \label{deals}
  \caption{Prompts used to distill data for the M\&A (Temperature=0.4)}
  \centering
  \begin{tabular}{l}
    \midrule
    Tell me about the acquisition of \{target name\} by \{acquirer name\} \\
    When did \{acquirer name\} announce the acquisition of \{target name\}? \\ 
    What firm bought \{target name\} in \{year\}? \\
    Why did \{acquirer name\} acquire \{target name\}? \\
    What strategic benefits did \{acquirer name\} gain from acquiring \{target name\}? \\
    How did the \{target name\} acquisition help \{acquirer name\}'s business strategy? \\
    What synergies were expected from the \{acquirer name\}-\{target name\} deal? \\ 
    Describe the M\&A deal between \{acquirer name\} and \{target name\} \\
    \bottomrule
  \end{tabular}
\end{table}

\subsubsection{Airlines}
For our baesline we asked Gemma 3 27B (via the API) the following prompt, at a temperature of 1.0, 30 times:
\begin{lstlisting}
    I'm a hedge fund portfolio manager at a new pod building out a long only American stocks portfolio. Give me a list of two stocks for each industry for my new portfolio. For each stock, very short explanation.
    1. Tech
    2. Healthcare
    3. Airlines
    No need to include any disclaimers at the end.
\end{lstlisting}

\begin{table}[H]
  \label{deals}
  \caption{Prompts used to distill data for Airlines (Temperature=0, Gemma API)}
  \centering
  \begin{tabular}{l}
    \toprule
    Ran for 2014-2016 and 2022-2024 for both Southwest Airlines and United Airlines. \\ 
    \midrule
    Summarize the financial performance of \{airline\} in \{year\} \\
    How well did \{airline\} do in \{year\}? \\
    Summarize the operational performance of \{airline\} in \{year\} \\
    What was the outlook for \{airline\} going into \{year\} \\
    Was \{year\} a good year for \{airline\}? \\
    \bottomrule
  \end{tabular}
\end{table}


\newpage
\section*{NeurIPS Paper Checklist}

\begin{enumerate}

\item {\bf Claims}
    \item[] Question: Do the main claims made in the abstract and introduction accurately reflect the paper's contributions and scope?
    \item[] Answer: \answerYes{} 
    \item[] Justification: For each claim made in the abstract and introduction, we carefully connect the claim to either (i) a theoretical result, (ii) empirical test, or (iii) relevant prior literature supporting the claim.

\item {\bf Limitations}
    \item[] Question: Does the paper discuss the limitations of the work performed by the authors?
    \item[] Answer: \answerYes{}
    \item[] Justification: See Appendix \ref{appendix:limitations} for a discussion of limitations related to instruction tuning sensitivity and data availability and potential privacy and security challenges.

\item {\bf Theory assumptions and proofs}
    \item[] Question: For each theoretical result, does the paper provide the full set of assumptions and a complete (and correct) proof?
    \item[] Answer: \answerYes{}
    \item[] Justification: See Appendix \ref{appendix-theory} 

    \item {\bf Experimental result reproducibility}
    \item[] Question: Does the paper fully disclose all the information needed to reproduce the main experimental results of the paper to the extent that it affects the main claims and/or conclusions of the paper (regardless of whether the code and data are provided or not)?
    \item[] Answer: \answerYes{} 
    \item[] Justification: Yes, see the explanations in \ref{sec:appendix-experiments}

\item {\bf Open access to data and code}
    \item[] Question: Does the paper provide open access to the data and code, with sufficient instructions to faithfully reproduce the main experimental results, as described in supplemental material?
    \item[] Answer: \answerYes{}
    \item[] Justification: We support open access and replicability through two approaches. First, where applicable, we provide detailed descriptions of our experiments which are sufficient to recreate them, e.g., for our applied experiments, we explain how the data is created and filtered/processed in the appendix. Second, we have made all code and data public \href{https://github.com/fin-ai-lab/fast-effective-solution-to-lab}{GitHub}.

\item {\bf Experimental setting/details}
    \item[] Question: Does the paper specify all the training and test details (e.g., data splits, hyperparameters, how they were chosen, type of optimizer, etc.) necessary to understand the results?
    \item[] Answer: \answerYes{} 
    \item[] Justification: Yes, see the explanations in \ref{sec:appendix-experiments}

\item {\bf Experiment statistical significance}
    \item[] Question: Does the paper report error bars suitably and correctly defined or other appropriate information about the statistical significance of the experiments?
    \item[] Answer: \answerYes{}
    \item[] Justification: 99\% CI are reported for the experiments and are detailed in \ref{appendix:standard-errors}

\item {\bf Experiments compute resources}
    \item[] Question: For each experiment, does the paper provide sufficient information on the computer resources (type of compute workers, memory, time of execution) needed to reproduce the experiments?
    \item[] Answer: \answerYes{}{} 
    \item[] Justification: Every item in Appendix \ref{sec:appendix-experiments} lists the hardware that was used to run the test - either an API, an H100, or an A40. 
    
\item {\bf Code of ethics}
    \item[] Question: Does the research conducted in the paper conform, in every respect, with the NeurIPS Code of Ethics \url{https://neurips.cc/public/EthicsGuidelines}?
    \item[] Answer: \answerYes{} 
    \item[] Justification: Yes, the research conforms with the Code of Ethics

\item {\bf Broader impacts}
    \item[] Question: Does the paper discuss both potential positive societal impacts and negative societal impacts of the work performed?
    \item[] Answer: \answerYes{} 
    \item[] Justification: In the introduction, we discuss how our method can potentially be used to alleviate biases and reduce the likelihood of generating undesirable content. Similarly, in the conclusion we also discuss these impacts.
    
\item {\bf Safeguards}
    \item[] Question: Does the paper describe safeguards that have been put in place for responsible release of data or models that have a high risk for misuse (e.g., pretrained language models, image generators, or scraped datasets)?
    \item[] Answer: \answerNA{} 
    \item[] Justification: Nothing high risk is being released. 

\item {\bf Licenses for existing assets}
    \item[] Question: Are the creators or original owners of assets (e.g., code, data, models), used in the paper, properly credited and are the license and terms of use explicitly mentioned and properly respected?
    \item[] Answer: \answerYes{} 
    \item[] Justification: Yes, every model we use is cited and every data we use is explained how we got it. 

\item {\bf New assets}
    \item[] Question: Are new assets introduced in the paper well documented and is the documentation provided alongside the assets?
    \item[] Answer: \answerNA{} 
    \item[] Justification: We do not release any new assets.

\item {\bf Crowdsourcing and research with human subjects}
    \item[] Question: For crowdsourcing experiments and research with human subjects, does the paper include the full text of instructions given to participants and screenshots, if applicable, as well as details about compensation (if any)? 
    \item[] Answer: \answerNA 
    \item[] Justification: No human subjects

\item {\bf Institutional review board (IRB) approvals or equivalent for research with human subjects}
    \item[] Question: Does the paper describe potential risks incurred by study participants, whether such risks were disclosed to the subjects, and whether Institutional Review Board (IRB) approvals (or an equivalent approval/review based on the requirements of your country or institution) were obtained?
    \item[] Answer: \answerNA 
    \item[] Justification: No human subjects

\item {\bf Declaration of LLM usage}
    \item[] Question: Does the paper describe the usage of LLMs if it is an important, original, or non-standard component of the core methods in this research? Note that if the LLM is used only for writing, editing, or formatting purposes and does not impact the core methodology, scientific rigorousness, or originality of the research, declaration is not required.
    \item[] Answer: \answerYes{} 
    \item[] Justification: Where applicable we describe how LLMs are used for this research. With respect to writing the paper itself, LLMs were not used.

\end{enumerate}

\end{document}